\documentclass[sigconf, natbib=true, dvipsnames]{acmart}

\usepackage{booktabs,multirow}
\usepackage{balance}
\usepackage{enumitem}
\usepackage{kotex}
\usepackage{color}
\usepackage{rotating}
\usepackage{amsmath}
\usepackage{array}
\usepackage[linesnumbered,algoruled,boxed,lined]{algorithm2e}

\newcommand{\ie}{{\it i.e.}}
\newcommand{\eg}{{\it e.g.}}

\newcommand{\ul}{\underline}{}

\usepackage[utf8]{inputenc}

\AtBeginDocument{%
  }

\copyrightyear{2023}
\acmYear{2023}
\setcopyright{acmlicensed}
\acmConference[CIKM '23] {Proceedings of the 32nd ACM International Conference on Information and Knowledge Management}{October 21--25, 2023}{Birmingham, United Kingdom.}
\acmBooktitle{Proceedings of the 32nd ACM International Conference on Information and Knowledge Management (CIKM '23), October 21--25, 2023, Birmingham, United Kingdom}
\acmPrice{15.00}
\acmISBN{979-8-4007-0124-5/23/10}
\acmDOI{10.1145/3583780.3615191}

\begin{document}

\title{Forgetting-aware Linear Bias for Attentive Knowledge Tracing}

\author{Yoonjin Im}\authornote{Equal contribution}
\affiliation{
  \institution{Sungkyunkwan University}
  \country{Suwon, Republic of Korea}}
\email{skewondr@skku.edu}

\author{Eunseong Choi}\authornotemark[1]
\affiliation{
  \institution{Sungkyunkwan University}
  \country{Suwon, Republic of Korea}}
\email{eunseong@skku.edu}

\author{Heejin Kook}
\affiliation{
  \institution{Sungkyunkwan University}
  \country{Suwon, Republic of Korea}}
\email{hjkook@skku.edu}

\author{Jongwuk Lee}\authornote{Corresponding author}
\affiliation{
  \institution{Sungkyunkwan University}
  \country{Suwon, Republic of Korea}}
\email{jongwuklee@skku.edu}

\renewcommand{\shortauthors}{Yoonjin Im, Eunseong Choi, Heejin Kook, \& Jongwuk Lee}

\begin{abstract}
Knowledge Tracing (KT) aims to track proficiency based on a question-solving history, allowing us to offer a streamlined curriculum. Recent studies actively utilize attention-based mechanisms to capture the correlation between questions and combine it with the learner's characteristics for responses. However, our empirical study shows that existing attention-based KT models neglect the learner's \emph{forgetting behavior}, especially as the interaction history becomes longer. This problem arises from the bias that overprioritizes the correlation of questions while inadvertently ignoring the impact of forgetting behavior. This paper proposes a simple-yet-effective solution, namely \emph{\textbf{Fo}rgetting-aware \textbf{Li}near \textbf{Bi}as (FoLiBi)}, to reflect forgetting behavior as a linear bias. Despite its simplicity, FoLiBi is readily equipped with existing attentive KT models by effectively decomposing question correlations with forgetting behavior. FoLiBi plugged with several KT models yields a consistent improvement of up to 2.58\% in AUC over state-of-the-art KT models on four benchmark datasets.
\end{abstract}


  \maketitle

\section{Introduction}\label{sec:introduction}

With the prevalence of various online education platforms and intelligent tutoring systems, analyzing the learner's proficiency through her interactions is crucial. \emph{Knowledge Tracing (KT)}~\cite{abs-2201-06953} aims to quantify the level of knowledge by tracking the learner's responses to the preceding questions and to predict the performance of new questions. To effectively solve this task as a sequential prediction, it is essential to learn the correlation between questions (\eg, question similarity and concept dependency) and the characteristics of the learner's response (\eg, forgetting behavior and concept proficiency).

Recent KT models utilized various deep neural networks to capture the question correlation: (i) Deep sequential models represent learner's knowledge with hidden states of RNNs~\cite{PiechBHGSGS15DKT, Guo21D1, Lee19D2, Liu21D3, Minn2018D4, Chen18D5, Su18D6, Yeung18D7}. (ii) Memory-augmented models~\cite{ZhangSKY17DKVMN, Abdel19M1} use key-value memory structures to update knowledge. (iii) Graph-based models employ GNNs to capture the correlation between questions and concepts~\cite{NakagawaIM19GKT, Tong20G1, Yang21G2}. (iv) Attentive models that apply several variants of attention mechanisms to reflect long-term relationships of questions~\cite{PandeyK19SAKT, LeeCLPP22CL4KT, ChoiLCBKCSBH20SAINT, LongQS0XTH022COKT, Shen20A1}.

Besides, several studies~\cite{NagataniZSCCO19DFKT, ShenLCHHYS021LPKT, Abdel2021DGMN, ChenLHWCWSH17KPT, ShinSYLKC21SAINT+, PandeyS20RKT, GhoshHL20AKT} attempted to consider \emph{forgetting behavior} based on a learning curve theory~\cite{Ebbinghaus2013forgetcurve} into deep learning models to integrate the learner's latent factors. A common assumption is that the learners' knowledge is increasingly forgotten over time. DFKT~\cite{NagataniZSCCO19DFKT} and LPKT~\cite{ShenLCHHYS021LPKT} extend DKT~\cite{PiechBHGSGS15DKT} by leveraging time-interval information to reflect the forgetting effect of learners. HawkesKT~\cite{WangMZLWLT0M21HawkesKT} utilizes time interval information to improve an existing matrix-factorization-based KT model~\cite{VieK19KTM}. SAINT+~\cite{ShinSYLKC21SAINT+} extends SAINT~\cite{ChoiLCBKCSBH20SAINT} by using the time gap between questions from different concept sets. However, these methods need help establishing discrete representations for continuous time intervals, mainly when they encounter time outliers that exceed the threshold of pre-defined embeddings. AKT~\cite{GhoshHL20AKT} and RKT~\cite{PandeyS20RKT} address this challenge by introducing a relative time interval bias that penalizes positional decay in attention weights instead of using absolute position embeddings. Nonetheless, the bias becomes unnecessarily entangled with question correlations, obscuring the influence of forgetting behavior. It is found that this tendency is more evident in longer interactions, suggesting the disentanglement between question correlations and forgetting behavior.


Inspired by recent work~\cite{PressSL22ALiBi}, we propose a more fine-designed method to model forgetting behavior built upon attention-based KT models, \emph{\textbf{Fo}rgetting-aware \textbf{Li}near \textbf{Bi}as (FoLiBi)}, decoupling the learner's forgetting effect from question correlations. It can be achieved through linear bias, which penalizes the question attention scores proportional to the relative distance. Despite its simplicity, it has two advantages. (i) It does not interfere with estimating the correlations of questions. (ii) It shows the robustness against the various sequence lengths. We have validated that FoLiBi consistently outperforms existing methods for modeling forgetting behavior on four standard datasets when plugged into four different KT models.

\begin{figure}
\includegraphics[width=0.80\linewidth]{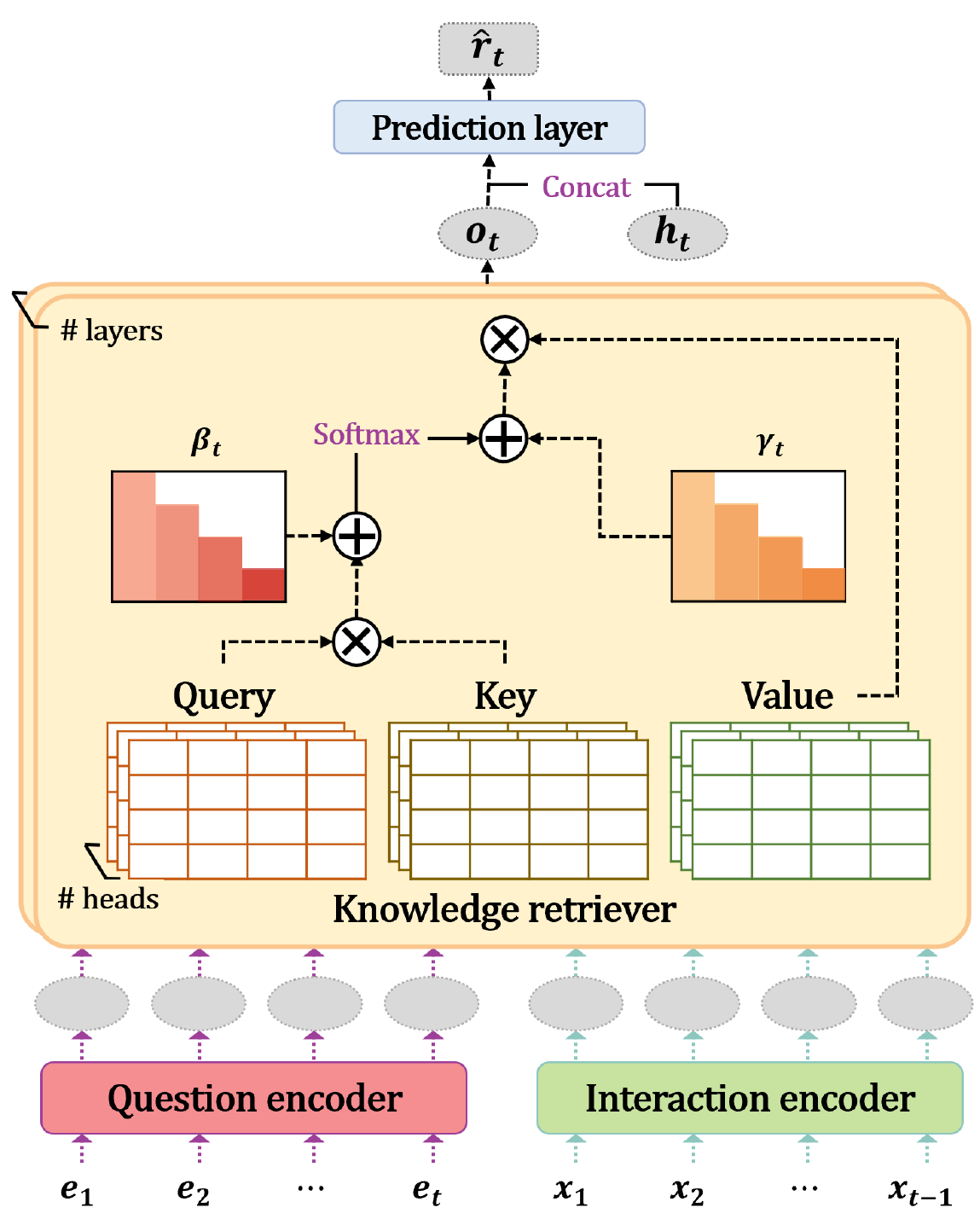}
\caption{The architecture of the forgetting-aware attentive KT model. We describe the proposed method in the knowledge retriever, a typical framework for attentive KT models.}\label{fig:fig1}
\vskip -0.2in
\end{figure}



\section{Proposed Method}\label{sec:method}
\textbf{Problem definition}. Let $X_{t} = \langle x_1, x_2 ..., x_{t}\rangle$ denote a sequence of interactions at the $t$-th time step. Each interaction $x_i=(e_i, r_i)$ consists of a pair of a question $e_i$ and its correctness $r_i \in \{0,1\}$. If $e_i$ is answered correctly, $r_i=1$, otherwise $r_i=0$. Given the sequence of interactions $X_{t-1}$, KT models aim to predict the probability of correctly answering the question $e_t$, \ie, $\hat{r}_t= p(r_t = 1 \mid X_{t-1}, e_t)$. They commonly adopt a binary cross-entropy for a loss function.



\vspace{1mm}
\noindent
\textbf{Attentive knowledge tracing models}.
Figure~\ref{fig:fig1} shows the architecture of attention-based KT models based on the previous works~\cite{GhoshHL20AKT, LeeCLPP22CL4KT}. Given sequences of questions $\langle e_1, e_2, ..., e_{t}\rangle$ and interactions $X_{t-1}$, we get the knowledge vector of the learner $\boldsymbol{o_t}\in \mathbb{R}^{1\times d}$ by passing through three attention mechanism-based structures: question encoder, interaction encoder, and knowledge retriever. It then combines $\boldsymbol{o_t}$ with the target question's embedding vector $\textbf{h}_t\in \mathbb{R}^{1\times d}$, and feeds it to a classification layer consisting of two-layered MLP with a sigmoid function to get a final prediction ${\hat{r}_t}$. 

While the two encoders and the knowledge retriever have similar internal architectures, our primary focus lies on the knowledge retriever. It calculates the question correlations as attention weights and aggregates interactions as a linear combination. 

\begin{equation}
\boldsymbol{\alpha}_t=\operatorname{softmax}\left(\frac{\textbf{q}_t \textbf{K}^{\top}}{\sqrt{d}} \right), \textbf{v}_t = \boldsymbol{\alpha}_t \textbf{V}, 
\end{equation}
where $d$ is the dimension size of key, query, and value embedding vectors. Let $\textbf{q}_t \in \mathbb{R}^{1 \times d}$, $\textbf{K} \in \mathbb{R}^{(t-1) \times d}$, and $\textbf{V} \in \mathbb{R}^{(t-1) \times d}$ denote the query vector for the target question $e_t$, the key matrix for previous questions $(1 \leq i < t)$, and the value matrix for previous responses, respectively. The scaled dot product \cite{VaswaniSPUJGKP17transformer} captures the relevance of each previous question with respect to the current questions. $\textbf{v}_t$ is computed as a weighted sum
of the values. 



\vspace{1mm}
\noindent
\textbf{Modeling forgetting behavior}. The correlation between a previous and the target question diminishes over time, indicating a gradual decay of the user's memory. We integrate the concept of forgetting behavior in AKT~\cite{GhoshHL20AKT} and RKT~\cite{PandeyS20RKT} into the attention blocks presented, resulting in a simplified formula. 
\begin{equation}
\label{eq:general_attention}
\boldsymbol{\alpha}_t=\operatorname{softmax}\left(\frac{\textbf{q}_t \textbf{K}^{\top}+\boldsymbol{\beta}_t}{\sqrt{d}}\right) + \boldsymbol{\gamma}_t,
\end{equation}
where $\boldsymbol{\beta}_{t} \in \mathbb{R}^{1\times (t-1)}$ and $\boldsymbol{\gamma}_{t} \in \mathbb{R}^{1\times (t-1)}$ are additional coefficients to represent forgetting behavior according to $t$-th time step. By combining these two additional terms into the attention mechanism, this approach considers both question correlations and forgetting behavior in a unified manner.

AKT~\cite{GhoshHL20AKT} uses \emph{monotonic attention (Mono)}, which incorporates forgetting behavior into the attention weight. Monotonic attention can generally be expressed with coefficient $\boldsymbol{\beta}_t$ as:
\begin{equation}
\label{eq:Mono}
\boldsymbol{\beta}_{t}^{\text {Mono }}=(\exp(-\theta_h \cdot \textbf{d}_{t})-1) \cdot \textbf{q}_{t} \textbf{K}^{\top}, \ \boldsymbol{\gamma}_t^{\text {Mono }}=\vec{0},
\end{equation}
where $\theta_h$ is a learnable decay rate parameter for the $h$-th head within the multi-head attention block. $\textbf{d}_{t}=[d(1, t), d(2, t), \ldots, d(t-1, t)]\in \mathbb{R}^{1\times (t-1)}$ consists of the function $d(i, t)$, which calculates the time interval bias among time steps $i$ and $t$ for a given question by multiplying the question similarity with relative distance. In other words, $\textbf{d}_{t}$ penalizes positional decay but unnecessarily intertwines with the correlation of the question.





Meanwhile, RKT~\cite{PandeyS20RKT} develops a \emph{relational coefficient (RC)} that considers question similarity and forgetting behavior. It can be defined with coefficient $\boldsymbol{\gamma}_t$ as:
\begin{equation}
\label{eq:RC}
\boldsymbol{\beta}_{t}^{\text {RC }}=\vec{0}, \ \boldsymbol{\gamma}_{t}^{\text {RC }}=\operatorname{softmax}(\textbf{R}^E+\textbf{R}^T),
\end{equation}
where $\textbf{R}^E\in \mathbb{R}^{1\times (t-1)}$ represents the similarities between previous questions and the target question derived from textual embeddings, and $\textbf{R}^{T}\in \mathbb{R}^{1\times (t-1)}$ is defined as $[\exp(-(t-1)/S), \exp(-(t-2)/S), \ldots, \exp(-1/S)]$, consisting of the relative time interval normalized by a learnable decay rate $S$. Because $\textbf{R}^E$ denotes the similarity of the context between the questions, the question correlation is again considered in the attention operation eventually, which reduces the effect of the relative time interval.



\begin{table}[t]
\caption{Dataset statistics.}
\label{tab:dataset}
\vspace{-3mm}
\centering
\label{tab:dataset}
\resizebox{\columnwidth}{!}{\begin{tabular}{c|ccccc} 
\toprule
Dataset & \# learners & \# interactions & \# concepts & \# questions & \% correct \\
\hline
AL05    & 571      & 607,014      & 271      & 173,113      & 76      \\
BR06    & 1,138    & 1,817,450    & 550      & 129,263      & 78      \\
AS09    & 3,695    & 282,071      & 149      & 17,728      & 62      \\
SLPY    & 5,000    & 623,212      & 1,380    & 2,713      & 79      \\
\bottomrule
\end{tabular}}
\vskip -0.2in
\end{table}
\begin{table*}[t!]\small
\caption{Overall performance comparisons. The best scores are in \textbf{bold}, and the second-best scores are \ul{underlined}. Statistical significant differences ($p<0.05$) between the reproduced baseline models and FoLiBi are reported with $^{\diamond}$. The last column shows the average ratio of the improvement over the original method, \ie, PE for SAKT~\cite{PandeyK19SAKT} and Mono for AKT~\cite{GhoshHL20AKT}, CL4KT~\cite{LeeCLPP22CL4KT}.}
\label{tab:performance}
\begin{tabular}{cc|ccc|ccc|ccc|ccc|c}
\toprule
\multirow{2}{*}{\textbf{Model}} & \multirow{2}{*}{\textbf{Type}} & \multicolumn{3}{c|}{\textbf{AL05}} & \multicolumn{3}{c|}{\textbf{BR06}} & \multicolumn{3}{c|}{\textbf{AS09}} & \multicolumn{3}{c|}{\textbf{SLPY}} & \multirow{2}{*}{\begin{tabular}[c]{@{}c@{}}AUC improv.\\ /Original\end{tabular}} \\
 &  & AUC & ACC & RMSE & AUC & ACC & RMSE & AUC & ACC & RMSE & AUC & ACC & RMSE &  \\ \hline
\multirow{4}{*}{\textbf{SAKT}} & PE & 75.69$^{\diamond}$ & 78.50$^{\diamond}$ & 39.37$^{\diamond}$ & 74.37$^{\diamond}$ & 78.83$^{\diamond}$ & 39.13$^{\diamond}$ & 73.81$^{\diamond}$ & 70.60$^{\diamond}$ & 44.42$^{\diamond}$ & 69.02$^{\diamond}$ & {\ul{77.94}} & 40.11 & \multirow{4}{*}{2.58\%} \\
 & Mono & 75.62$^{\diamond}$ & 78.57$^{\diamond}$ & 39.20$^{\diamond}$ & 74.43$^{\diamond}$ & 79.01$^{\diamond}$ & 38.77$^{\diamond}$ & 73.62$^{\diamond}$ & 70.38$^{\diamond}$ & 44.35$^{\diamond}$ & 69.35$^{\diamond}$ & \textbf{78.21} & \textbf{39.64} &  \\
 & RC & {\ul{76.31}$^{\diamond}$} & {\ul{78.68}$^{\diamond}$} & {\ul{39.09}$^{\diamond}$} & {\ul{75.42}$^{\diamond}$} & {\ul{79.15}$^{\diamond}$} & {\ul{38.71}$^{\diamond}$} & {\ul{74.12}$^{\diamond}$} & {\ul{70.90}$^{\diamond}$} & {\ul{44.23}$^{\diamond}$} & {\ul{69.62}$^{\diamond}$} & 77.35 & 40.18 &  \\
 & FoLiBi & \textbf{77.96} & \textbf{79.40} & \textbf{38.57} & \textbf{76.72} & \textbf{79.69} & \textbf{38.25} & \textbf{75.24} & \textbf{71.78} & \textbf{43.68} & \textbf{70.55} & 77.52 & {\ul{40.01}} &  \\ \hline
\multirow{4}{*}{\textbf{AKT}} & PE & 74.13$^{\diamond}$ & 77.68$^{\diamond}$ & 39.88$^{\diamond}$ & 74.64$^{\diamond}$ & 78.47$^{\diamond}$ & 38.86$^{\diamond}$ & 73.49$^{\diamond}$ & 70.27$^{\diamond}$ & 44.59$^{\diamond}$ & 70.48$^{\diamond}$ & 78.06 & 39.66$^{\diamond}$ & \multirow{4}{*}{0.84\%} \\
 & Mono & {\ul{76.73}$^{\diamond}$} & 78.61 & 38.93$^{\diamond}$ & 76.18$^{\diamond}$ & {\ul{79.31}} & 38.26$^{\diamond}$ & {\ul{74.91}$^{\diamond}$} & {\ul{71.25}$^{\diamond}$} & {\ul{43.80}} & {\ul{72.42}$^{\diamond}$} & 78.49 & {\ul{39.03}} &  \\
 & RC & 76.63$^{\diamond}$ & {\ul{78.62}} & {\ul{38.89}$^{\diamond}$} & {\ul{76.42}$^{\diamond}$} & 79.25 & {\ul{38.22}$^{\diamond}$} & 74.81$^{\diamond}$ & 71.13$^{\diamond}$ & 43.92$^{\diamond}$ & 72.15$^{\diamond}$ & \textbf{78.71} & 39.13$^{\diamond}$ &  \\
 & FoLiBi & \textbf{77.64} & \textbf{78.89} & \textbf{38.64} & \textbf{77.09} & \textbf{79.39} & \textbf{38.09} & \textbf{75.28} & \textbf{71.60} & \textbf{43.63} & \textbf{72.78} & {\ul{78.69}} & \textbf{38.90} &  \\ \hline
\multirow{4}{*}{\textbf{CL4KT}} & PE & 74.93$^{\diamond}$ & 77.76$^{\diamond}$ & 39.64$^{\diamond}$ & 74.02$^{\diamond}$ & 78.39$^{\diamond}$ & 39.16$^{\diamond}$ & 74.00$^{\diamond}$ & 70.48$^{\diamond}$ & 44.39$^{\diamond}$ & 68.99$^{\diamond}$ & 78.18 & 40.01$^{\diamond}$ & \multirow{4}{*}{1.13\%} \\
 & Mono & {\ul{77.72}$^{\diamond}$} & {\ul{79.09}$^{\diamond}$} & {\ul{38.52}$^{\diamond}$} & {\ul{76.65}$^{\diamond}$} & 79.55$^{\diamond}$ & 38.10$^{\diamond}$ & {\ul{75.47}$^{\diamond}$} & {\ul{71.54}$^{\diamond}$} & {\ul{43.56}$^{\diamond}$} & {\ul{71.27}} & 78.57 & 39.25 &  \\
 & RC & 77.03$^{\diamond}$ & 78.94$^{\diamond}$ & 38.71$^{\diamond}$ & 76.58$^{\diamond}$ & {\ul{79.70}} & {\ul{38.08}$^{\diamond}$} & 75.35$^{\diamond}$ & 71.36$^{\diamond}$ & 43.70$^{\diamond}$ & 71.23 & \textbf{78.71} & {\ul{39.23}} &  \\
 & FoLiBi & \textbf{78.97} & \textbf{79.69} & \textbf{38.09} & \textbf{77.80} & \textbf{80.00} & \textbf{37.73} & \textbf{76.20} & \textbf{72.00} & \textbf{43.29} & \textbf{71.59} & {\ul{78.69}} & \textbf{39.19} &  \\ \hline
\multirow{4}{*}{\shortstack{\textbf{CL4KT}\\ w/o CL}} & PE & 75.68$^{\diamond}$ & 78.16$^{\diamond}$ & 39.30$^{\diamond}$ & 75.78$^{\diamond}$ & 79.16$^{\diamond}$ & 38.54$^{\diamond}$ & 74.31$^{\diamond}$ & 70.73$^{\diamond}$ & 44.23$^{\diamond}$ & 71.04$^{\diamond}$ & 78.25$^{\diamond}$ & 39.56$^{\diamond}$ & \multirow{4}{*}{1.54\%} \\
 & Mono & {\ul{77.72}$^{\diamond}$} & {\ul{79.25}$^{\diamond}$} & {\ul{38.45}$^{\diamond}$} & 76.96$^{\diamond}$ & 79.68$^{\diamond}$ & 37.94$^{\diamond}$ & 75.18$^{\diamond}$ & 71.35$^{\diamond}$ & 43.72$^{\diamond}$ & 72.02$^{\diamond}$ & 78.75 & 39.14$^{\diamond}$ &  \\
 & RC & 77.54$^{\diamond}$ & 79.24$^{\diamond}$ & 38.49$^{\diamond}$ & {\ul{77.31}$^{\diamond}$} & {\ul{79.99}$^{\diamond}$} & {\ul{37.78}$^{\diamond}$} & {\ul{75.49}$^{\diamond}$} & {\ul{71.57}$^{\diamond}$} & {\ul{43.57}$^{\diamond}$} & {\ul{72.15}$^{\diamond}$} & {\ul{78.76}} & {\ul{38.97}$^{\diamond}$} &  \\
 & FoLiBi & \textbf{79.24} & \textbf{79.89} & \textbf{37.89} & \textbf{78.23} & \textbf{80.22} & \textbf{37.53} & \textbf{76.17} & \textbf{72.07} & \textbf{43.32} & \textbf{72.90} & \textbf{78.80} & \textbf{38.76} &  \\
\bottomrule
\end{tabular}
\end{table*}

Despite diligent efforts, Eq.~\eqref{eq:Mono} and~\eqref{eq:RC} are insufficient in capturing the exclusive influence of relative positional decay, resulting in only modest enhancements (see Section \ref{sec:results} for further discussion). We analyzed that both methods do not account for position decay in a balanced manner due to unnecessary redundancy. 



The linear bias~\cite{PressSL22ALiBi} computes a constant value based on relative distance and provides an effective approach to penalize forgetting behavior directly. This paper devises a simple-yet-effective solution, called \emph{forgetting-aware linear bias (FoLiBi)}, to decouple the intricate relationship between question correlation and forgetting behavior. We represent forgetting behavior using coefficient $\boldsymbol{\beta}_t$ as:
\begin{equation}
\boldsymbol{\beta}_{t}^{\text {FoLiBi}}=m_h\cdot [1, \dots, t-1], \ \boldsymbol{\gamma}_{t}^{\text {FoLiBi}}=\vec{0},
\end{equation}
where $[1, \dots, t-1]$ is a linear bias over time and $m_h=2^{\frac{-8}{H}\cdot h}$ denotes a coefficient that adjusts the importance of forgetting behavior for the $h$-th head out of $H$ attention heads.

Figure~\ref{fig:fig_RQ2} shows how the attention weights of the existing methods and FoLiBi are distributed according to the positions. Because FoLiBi independently computes the question correlations, it effectively penalizes positional decay for forgetting behavior, especially as the length of history increases, \eg, 20. Our experiments have shown that this change improves overall performance, suggesting that FoLiBi not only penalizes position collapse but also balances correlations between questions at the same time.


\section{Experiments}\label{sec:experiments}

\subsection{Experimental Setup}\label{sec:dataset}
\vspace{1mm}
\noindent
\textbf{Datasets.}
We evaluate the effectiveness of the proposed method using four real-world educational datasets: Algebra 2005-2006 (AL05), Bridge to Algebra 2006-2007 (BR06), ASSISTments 2009-2010 (AS09) \cite{FengHK09AS09}, and Slepemapy (SLPY) \cite{PapousekPS16SLPY}. AL05 and BR06 are algebra learning history datasets provided by KDD Cup 2010. AS09 is collected from a free online math education platform. SLPY, a dataset dedicated to geography learning history, originates from the online platform \textit{slepemapy.cz}, and we randomly sampled 5,000 learners out of 91,331. For all datasets, we follow the same pre-processing procedure in \cite{LeeCLPP22CL4KT, Gervet20Preprocess}, such as dropping learners whose number of interactions is less than 5. Table~\ref{tab:dataset} shows statistics of the processed datasets.

\vspace{1mm}
\noindent
\textbf{Baselines and evaluation metrics.}
We apply FoLiBi and other forgetting behavior modeling methods to several transformer-based models to analyze their effectiveness thoroughly. SAKT~\cite{PandeyK19SAKT} is an early model of knowledge tracing using an attention mechanism. AKT~\cite{GhoshHL20AKT} introduces embeddings based on the Rasch model~\cite{Rasch93Rasch} and reflects the learner's forgetting behavior in the attention mechanism. CL4KT~\cite{LeeCLPP22CL4KT} proposes contrastive learning on the augmented learning histories. We only report the method used in RKT~\cite{PandeyS20RKT}, \ie, RC, because it integrates textual content into the input in the backbone model. Following the previous works~\cite{LeeCLPP22CL4KT, ShenLCHHYS021LPKT}, we adopt Area Under Curve (AUC), Accuracy (ACC), and Root Mean Squared Error (RMSE), which are commonly used in KT for evaluation.
%

\vspace{1mm}
\noindent
\textbf{Implementation details.} We perform 5-fold cross-validation and report the average of 3 different seeds. For validation, we split 10\% from each training fold. Following ~\cite{GhoshHL20AKT, LeeCLPP22CL4KT}, we regard the same concepts as a single question to help avoid over-parameterization. We set the maximum history length to 100 for all experiments except the one for figure \ref{fig:fig_RQ3}. The hyperparameters are optimized for each dataset using Optuna~\cite{AkibaSYOK19OPTUNA}. All the source code is available.\footnote{\url{https://github.com/skewondr/FoLiBi}}

\subsection{Results and Analysis}\label{sec:results}

To verify the effectiveness of our proposed method, we address the following research questions.

\begin{itemize}[leftmargin=5mm]
\item \textbf{RQ1} Does FoLiBi improve the performance of the state-of-the-at KT models?
\item \textbf{RQ2} What are the main differences between FoLiBi and existing methods on attention weights?
\item \textbf{RQ3} Does FoLiBi perform better on longer histories?
\end{itemize}

\noindent
\textbf{Overall performance comparison}. Table~\ref{tab:performance} shows the performance of different forgetting behavior modeling methods using different backbone models on the four benchmark datasets. While Positional Embedding (PE) incorporates absolute position information into the input embedding, Monotonic attention (Mono), Relational Coefficient (RC), and FoLiBi design the coefficients $\boldsymbol{\beta_t}$ and $\boldsymbol{\gamma_t}$ to reflect relative positional decay. The key observations are: (i) All backbone models show the highest AUC performance when FoLiBi is plugged in. It confirms that FoLiBi helps trace learners' knowledge effectively by avoiding the intervention between two objectives, \ie, question and positional correlation. (ii) Mono and RC usually perform better than PE by accounting for relative distance. However, in most cases, they have no significant improvement because both approaches neglect the negative effects of the question correlation when considering forgetting behavior. (iii) Applying FoLiBi to CL4KT~\cite{LeeCLPP22CL4KT} shows an improvement over the original method, Mono, with a 1.13\% in AUC on average across datasets, but removing contrastive learning (CL) increases the improvement to 1.54\%. As the contrastive learning in CL4KT~\cite{LeeCLPP22CL4KT} leads the model to be more dependent on the question relevance, it implies that the balance between the question and position correlation is crucial.

\begin{figure}
\includegraphics[width=0.97\linewidth]{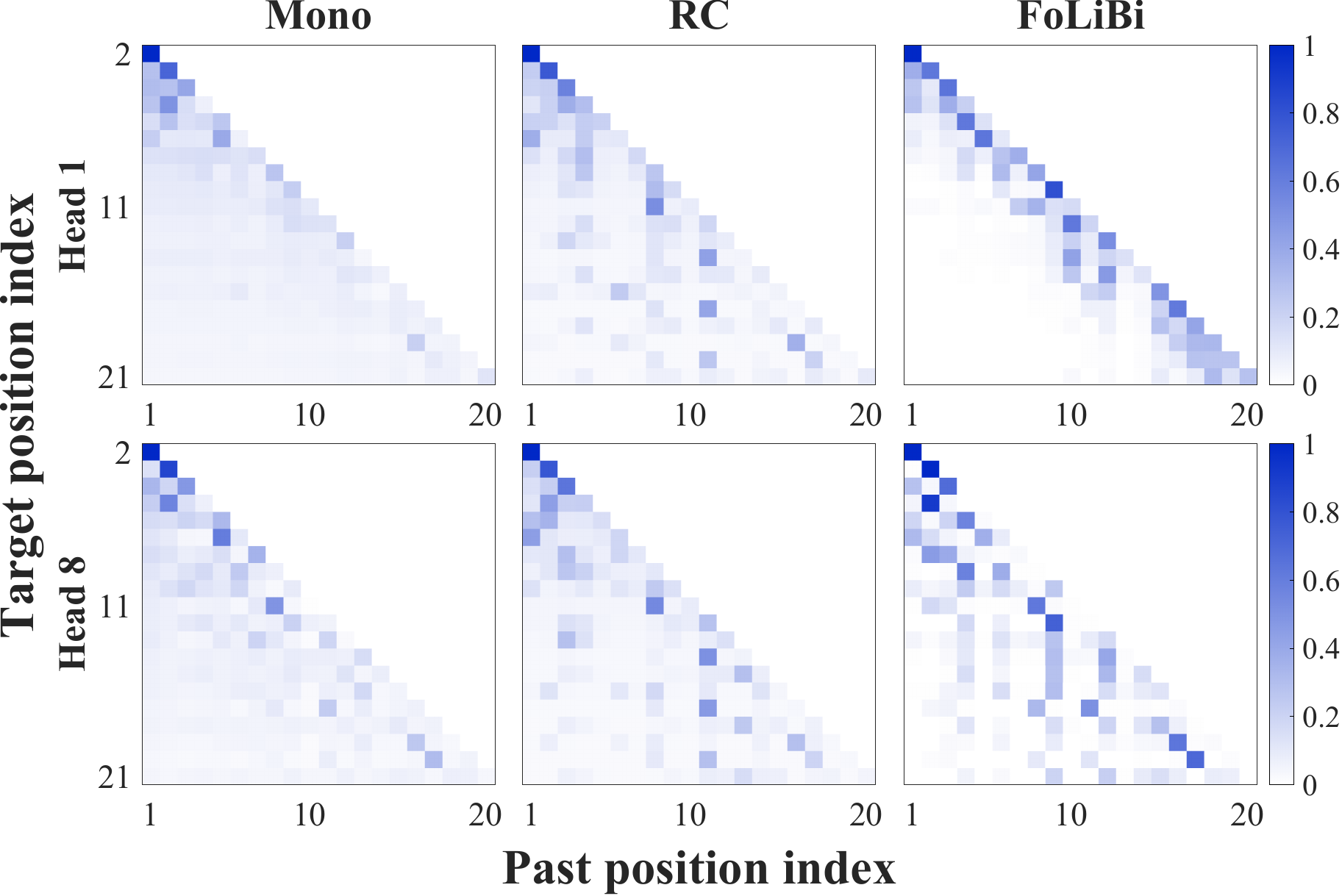}
\vskip -0.05in
\caption{Visualization of attention weights with different methods for forgetting behavior on the AKT. Note that the value in the upper triangle is always 0 due to the causal mask.}\label{fig:fig_RQ2}
\vskip -0.1in
\end{figure}

\begin{figure}[t]
\includegraphics[width=1.0\linewidth]{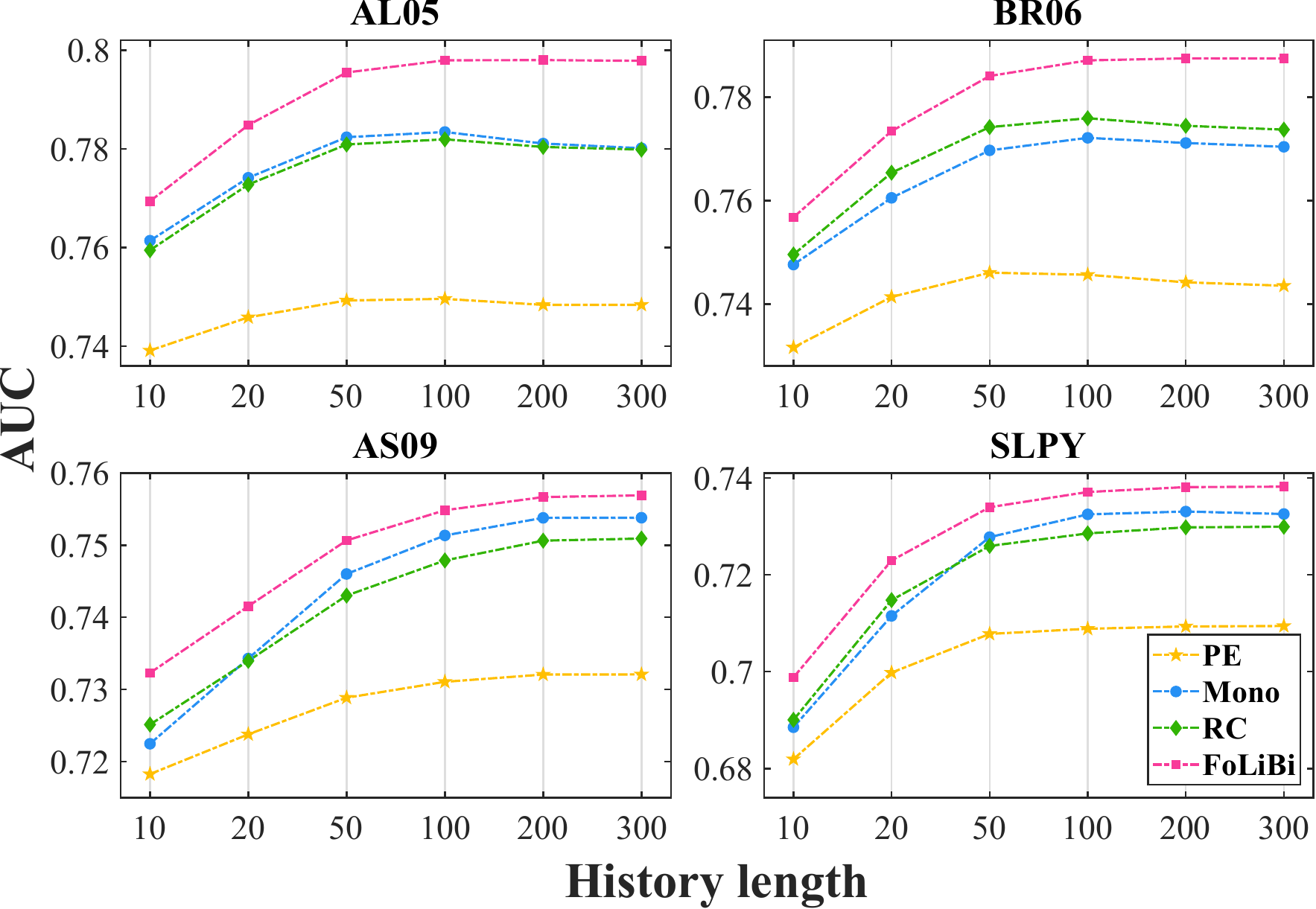}
\vskip -0.1in
\caption{Performance comparison varying the number of previous interactions used as input for inference. We used the AKT model as the backbone and trained with a maximum length of 300 for a fair comparison to the PE method.}
\label{fig:fig_RQ3}
\vskip -0.11in
\end{figure}

\vspace{1mm}
\noindent
\textbf{Comparison via attention weights.}
Figure~\ref{fig:fig_RQ2} visualizes the attention weights of the first 20 records in the first and last head of the final self-attention block in AKT~\cite{GhoshHL20AKT}. The x and y-axis of each map represent the previous and the target questions, respectively. Mono and RC models tend to assign a weight spread across all previous questions in the first head. On the other hand, we found that FoLiBi holds the more relative position for rearward sequences. This is because Mono and RC are overly affected by question correlation when computing $\boldsymbol{\beta}_t$ and $\boldsymbol{\gamma}_t$. Nevertheless, FoLiBi uses a non-learnable $\boldsymbol{\beta}_t$, which means that a positional correlation can constantly influence the attention weight regardless of the model performance. While the other two methods maintain the tendency within heads, FoLiBi adjusts the importance of position correlation with the slope $m_h$, so that the last head assigns attention weights based mainly on the question correlations. This allows FoLiBi to capture complex interactions, resulting in consistent effectiveness over varying lengths. We will discuss this in the next section.

\vspace{1mm}
\noindent
\textbf{Effectiveness on various history lengths.}
Assuming the importance of forgetting behavior depends on the length of the problem-solving history, we conducted further experiments to see how effective each method is in varying lengths. We set the maximum sequence length to 300 in training and fixed the length of input sequence $n$, \ie, 10, 20, 50, 100, 200, or 300, in evaluation. For example, if $L$ is the total length of a learner's history, then the number of questions evaluated for the learner is $L-n$. We make the following observations. Firstly, the lines of RC, Mono, and FoLiBi are higher than those of PE, and the gaps between them get larger as the number of interactions increases. This suggests that relative positional distance is a core factor in the KT task and becomes more critical with a longer history. Second, the improvement of FoLiBi is maintained or increased as the number of interactions increases. It indicates that FoLiBi can reflect forgetting behavior better than other methods owing to decoupling from question correlations.

\section{Conclusion}\label{sec:conclusion}


This paper first analyzes the effect of forgetting behavior in existing attention-based KT models. As the interaction sequence extends over time, the correlation between questions becomes disproportionately emphasized compared to the impact of forgetting behavior. To address this problem, we propose FoLiBi, which represents forgetting behavior as a linear bias decoupled from the question correlation. Despite its simplicity, extensive experiments show that FoLiBi consistently outperforms previous KT models and distinguishes the relative position distance from the question correlation. As a result, FoLiBi is suitable to account for forgetting-aware attention by readily plugging into existing attention-based KT models.

\section*{Acknowledgments} 

This work was supported by Institute of Information \& communications Technology Planning \& Evaluation (IITP) grant and National Research Foundation of Korea (NRF) grant funded by the Korea government (MSIT) (No. 2019-0-00421, 2022-0-00680, 2022-0-01045, 2021-0-01560, and NRF-2018R1A5A1060031).

\newpage
\bibliographystyle{ACM-Reference-Format}
\balance
\bibliography{references}

\end{document}